\pdfoutput=1
\documentclass[11pt]{article}
\usepackage{emnlp2021}
\usepackage{times}
\usepackage{latexsym}

\usepackage{adjustbox} 
\usepackage{amssymb} 
\usepackage{microtype}
\usepackage{booktabs}

\usepackage{color}
\usepackage{changes}

\definechangesauthor[name=Andreas, color=orange]{andreas}

\definechangesauthor[name=Ekta, color=blue]{ekta}

\definechangesauthor[name=Dominike, color=brown]{dominike}

\definechangesauthor[name=Philipp, color=green]{philipp}

\definechangesauthor[name=Mihai, color=red]{mihai}

\definecolor{applegreen}{rgb}{0.55, 0.71, 0.0}

\usepackage{xspace} 
\newcommand{\methodName}{MULAN\xspace} 
\newcommand{\longMethodName}{Multimodal Human-like Attention Network\xspace} 

\title{Multimodal Integration of Human-Like Attention\\in Visual Question Answering}

\author{Ekta Sood$^1$, Fabian Kögel$^1$, Philipp Müller$^2$, Dominike Thomas,$^1$, Mihai Bâce$^1$, Andreas Bulling$^1$\\
$^1$University of Stuttgart, Institute for Visualization and Interactive Systems (VIS), Germany\\ $^2$German Research Center for Artificial Intelligence (DFKI)\\
\texttt{\small\{ekta.sood,fabian.koegel,dominike.thomas,mihai.bace,andreas.bulling\}@vis.uni-stuttgart.de} \\
\texttt{\small\{philipp.mueller\}@dfki.de}
}

\date{}

\begin{document}
\maketitle
\begin{abstract}
Human-like attention as a supervisory signal to guide neural attention has shown significant promise but is currently limited to unimodal integration -- even for inherently multimodal tasks such as visual question answering (VQA). 
We present the \longMethodName (\methodName) -- the first method for multimodal integration of human-like attention on image and text during training of VQA models.
\methodName integrates attention predictions from two state-of-the-art text and image saliency models into neural self-attention layers of a recent transformer-based VQA model.
Through evaluations on the challenging VQAv2 dataset, we show that \methodName achieves a new state-of-the-art performance of 73.98\% accuracy on test-std and 73.72\% on test-dev and, at the same time, has approximately 80\% fewer trainable parameters than prior work.
Overall, our work underlines the potential of integrating multimodal human-like and neural attention for VQA.
\end{abstract}

\section{Introduction}
Visual question answering (VQA) is an important task at the intersection of natural language processing (NLP) and computer vision that has attracted significant research interest in recent years
~\cite{antol_vqa_2015, malinowski2014-amultiworld, malinowski2015-askyourneurons,yu_deep_2019}. One of its key challenges is, by its very nature, to jointly analyze and understand the language and visual input computationally. State-of-the-art methods for VQA rely on neural attention mechanisms to encode relations between questions and images and, for example, focus processing on parts of the image that are particularly relevant for a given question~\cite{yu2017multi,yu_deep_2019,jiang_defense_2020,anderson_bottom-up_2018}. An increasing number of methods make use of multiple Transformer-based attention modules~\cite{Vaswani.2017}: they enable attention-based text features to exert complex patterns of influence on attention processes on  images~\cite{yu_deep_2019,jiang_defense_2020}. 

Simultaneously, an increasing number of works have demonstrated the effectiveness of integrating human-like attention into neural attention mechanisms across a wide range of tasks, including image captioning~\cite{sugano2016seeing}, text comprehension tasks~\cite{sood_improving_2020}, or sentiment analysis and grammatical error detection~\cite{barrett2018sequence}. Research on integrating human-like attention into neural attention mechanisms for VQA has also attracted increasing interest~\cite{qiao_exploring_nodate,Selvaraju.2019, Wu.2019, chen_air_2020, Shrestha.12.04.2020}. However, despite the inherent multimodality of the VQA task, these works only integrate human-like attention on images.
A method for predicting and integrating human-like attention on text, which has obtained state-of-the-art performance in downstream NLP tasks, has only recently been proposed~\cite{sood_improving_2020} and multimodal integration of human-like attention remains unexplored.

We fill this gap by proposing the \longMethodName (\methodName) -- the first method for multimodal integration of human-like attention in VQA.
In contrast to previous unimodal integration methods on images alone, our method allows human attention information to act as a link between text and image input.
\methodName integrates state-of-the-art human saliency models for text and images into the attention scoring functions of the self-attention layers of the MCAN VQA architecture~\cite{yu_deep_2019}.
This way, human attention acts as an inductive bias directly modifying neural attention processes.
To model human-like attention on text we make use of the recently proposed Text Saliency Model (TSM)~\cite{sood_improving_2020} that we adapt to the VQA task while training the MCAN framework. 
Human attention on images is integrated using the recent multi-duration saliency model~\cite{fosco2020much} that also models the temporal dynamics of attention. We train our model on the VQAv2 dataset~\cite{Goyal.2017} and achieve state-of-the-art performance of 73.98\% accuracy on \emph{test-std} and 73.72\% \emph{test-dev}.
Notably, given that our model is based on the MCAN small variant, we require significantly fewer trainable parameters than the large variant.

Our contributions are three-fold: 
First, we propose a novel method to jointly integrate human-like attention on text and image into the MCAN VQA framework~\cite{yu_deep_2019}.
Second, we evaluate our method on the challenging \emph{test-std} VQAv2 benchmark~\cite{Goyal.2017} and
show that it
outperforms the state of the art on both \emph{test-std} and \emph{test-dev} while requiring about 80\% fewer trainable parameters.
Finally, through detailed analysis of success and failure cases we provide insights into how \methodName makes use of human attention information to correctly answer
questions that are notoriously difficult, e.g. longer questions. 
\section{Related Work}

Our work is related to previous works on 1) visual question answering, 2) using neural attention mechanisms, and 3) using human-like attention as a supervisory signal.

\paragraph{Visual Question Answering.}
Using natural language to answer a question based on a single image \citep{antol_vqa_2015} has been a topic of increasing interest in recent years~\cite{malinowski2014-amultiworld, malinowski2015-askyourneurons}.
\citet{antol_vqa_2015} built the first, large-scale VQA dataset
that provided open-ended, free-form questions created by humans.
Given that models have been shown to exploit bias in datasets~\citep{agrawal_analyzing_2016}, \citet{Goyal.2017}  expanded the VQA dataset by balancing it so that each question had two images, with two different answers to the same question. 
Tests on this new dataset (VQAv2) obtained significantly reduced performance for current models, showing how prevalent answer bias was before.
Another challenge in VQA remains the lack of inconsistency in answer predictions~\citep{shrestha_answer_2019, zhang_yin_2016, selvaraju_squinting_2020} and reduced performance for compositional questions \citep{agrawal_c-vqa_2017,andreas_neural_2016,shrestha_answer_2019} or linguistic variation~\citep{shah_cycle-consistency_2019, Agrawal.2018}.

\paragraph{Neural Attention Mechanisms.}
To imbue models with more reasoning capabilities,
researchers started experimenting with human-inspired neural attention and showed that adding neural attention mechanisms improved performance for VQA.
\citet{shih_where_nodate} added a region selection layer to pinpoint relevant areas of an image and improved from \citet{antol_vqa_2015} by 5\%.
Similarly, \citet{anderson_bottom-up_2018} demonstrated that using bottom-up attention was preferable to top-down, and won first place in the 2017 VQA Challenge. \citet{jiang_pythia_2018} further expanded on this work by optimizing the model architecture and won the 2018 VQA challenge.
Follow-up works combined learned visual and language attention in order to narrow down which part of the image and question are relevant, first with alternating attention \citep{lu_hierarchical_2017}, dual attention \citep{nam_dual_2017}, and finally multi-level attention \citep{yu2017multi}.
The success of Transformers \cite{Vaswani.2017} in NLP tasks also inspired new work in VQA. 
\citet{yu_deep_2019} created the Transformer-inspired Modular Co-Attention Network (MCAN) that combines self-attention with guided-attention to leverage the interaction within and between modalities.
\citet{jiang_defense_2020} further built on this architecture and won the 2020 VQA Challenge. 
\citet{tan_lxmert_2019} improved input encoding with a Transformer and transfer learning,
while \citet{li_oscar_2020} modified input encodings by adding an object tag to help align images and text semantically.

\paragraph{Supervision Using Human Attention.} 
Despite its advantages,
it was also demonstrated that neural attention may focus on the wrong area of an image~\citep{das_human_2016, chen_air_2020}.
To rectify this, human attention was brought in as an additional supervisory signal.
Researchers investigated differences between neural and human attention in VQA~\citep{das_human_2016,chen_air_2020} and created datasets containing human attention maps~\citep{fosco2020much, chen_air_2020, das_human_2016}.
At the same time, integrating human attention supervision showed to be promising in closely related computer vision~\citep{sugano2016seeing, karessli_gaze_2017} or NLP tasks~\citep{barrett2018sequence, sood20_conll}.
\citet{sood_improving_2020} proposed a novel text saliency model that, by combining a cognitive model of reading with human attention supervision, beat state-of-the-art results on paraphrase generation and sentence compression.
For VQA tasks, \citet{Gan.2017} combined human attention on images and semantic segmentation of questions. Using ground truth human attention, \citet{Wu.2019} penalized networks for focusing on the wrong area of an image, while \citet{Selvaraju.2019} guided neural networks to look at areas of an image that humans judged as particularly relevant for question answering. \citet{chen_air_2020} continued in this direction by using human attention to encourage reasoning behaviour from a model. Since obtaining ground truth human attention annotations is costly and time-consuming, \citet{qiao_exploring_nodate} trained a network on the VQA-HAT dataset to automatically generate human-like attention on unseen images, then used these saliency maps to create the enhanced Human-Like ATention (HLAT) dataset. 
\section{Method}

\begin{figure*}[ht]
  \includegraphics[width=\textwidth]{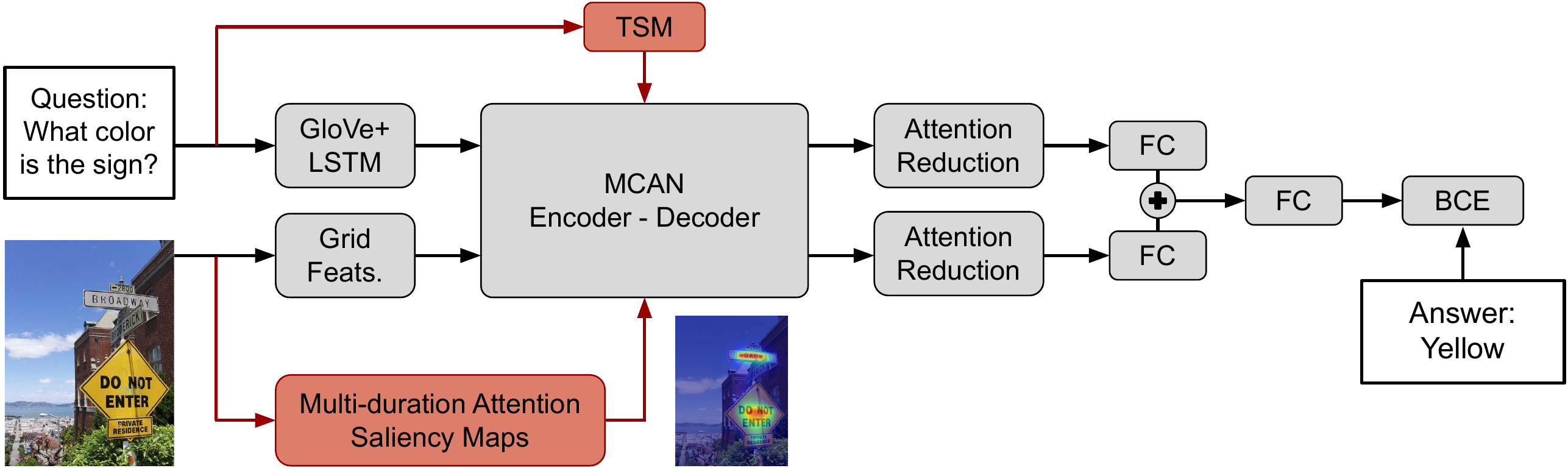}
  \caption{Overview of the Multimodal Human-like Attention Network (\methodName). Our method proposes multimodal integration of human-like attention on questions as well as images during training of VQA models. \methodName leverages attention predictions from two state-of-the-art text~\cite{sood_improving_2020} and image saliency models~\cite{fosco2020much}.}
  \label{fig:overall_arch}
\end{figure*}

The core contribution of our work is to propose \methodName, the first multimodal method to integrate human-like attention on both the image and text for VQA (see \autoref{fig:overall_arch} for an overview of the method).
At its core, our method builds on the recent MCAN model~\cite{yu_deep_2019,yu2019openvqa}, which won the 2019 VQA challenge as well as being the basis of the 2020 winning method utilizing grid features \cite{jiang_defense_2020}. We adapted the open source implementation\footnote{\url{ https://github.com/MILVLG/openvqa}} and trained the small variant of the MCAN using the grid features\footnote{\url{https://github.com/facebookresearch/grid-feats-vqa}}.
We first introduce the feature representations and the MCAN framework and subsequently explain our novel multimodal integration method\footnote{Our implementation and other supporting material will be made publicly available}.

\paragraph{Feature Representations.}
We represent the input images by spatial grid features, following the extraction methodology of \citet{jiang_defense_2020}. A Faster R-CNN with ResNet-50 backbone ~\cite{ren15fasterrcnn, He2015} is pre-trained on ImageNet ~\cite{deng2009imagenet} and VG~\cite{krishnavisualgenome} and then the object proposal and RoI pooling (used for region features in \citet{anderson_bottom-up_2018}) is removed. The remaining ResNet directly outputs the grid features. 
We obtain up to 19x32 features (depending on aspect ratio) per image. The final image representation is $X \in \mathbb{R}^{m\times d_x}$ with $m \in [192, 608]$, where $m$ represents the number of features, and $d_x$ the feature embedding dimension. 

The input questions are represented as in MCAN: tokenized at word-level, trimmed to $n\in[1, 14]$ tokens and represented using 300-D GloVe \cite{pennington2014glove} word embeddings. The $n\times300$ embeddings are further passed through a one-layer LSTM with hidden size $d_y$ and all intermediate hidden states form the final question representation matrix $Y\in\mathbb{R}^{n\times d_y}$.
Both representations are zero-padded to accommodate the varying number of grid features and question words.

\paragraph{Base model.}
In general, an attention function computes an alignment score between a query and key-value pairs and uses the score to re-weight the values. 
Attention methods differ in their choice of scoring function, whether they attend to the whole (global/soft) or only parts (local/hard) of the input, and whether queries and key-value pairs are projections from the same inputs (self-attention) or different inputs (guided attention). 
The Deep Modular Co-Attention Network (MCAN) for VQA~\cite{yu_deep_2019} is a Transformer-based network~\cite{Vaswani.2017} that runs multiple layers of multi-headed self-attention (SA) and guided-attention (GA) modules in an encoder-decoder architecture using the scaled dot-product score function.

\begin{figure}[t]
  \centering
  \hspace{-1cm}
  \includegraphics[width=.75\columnwidth]{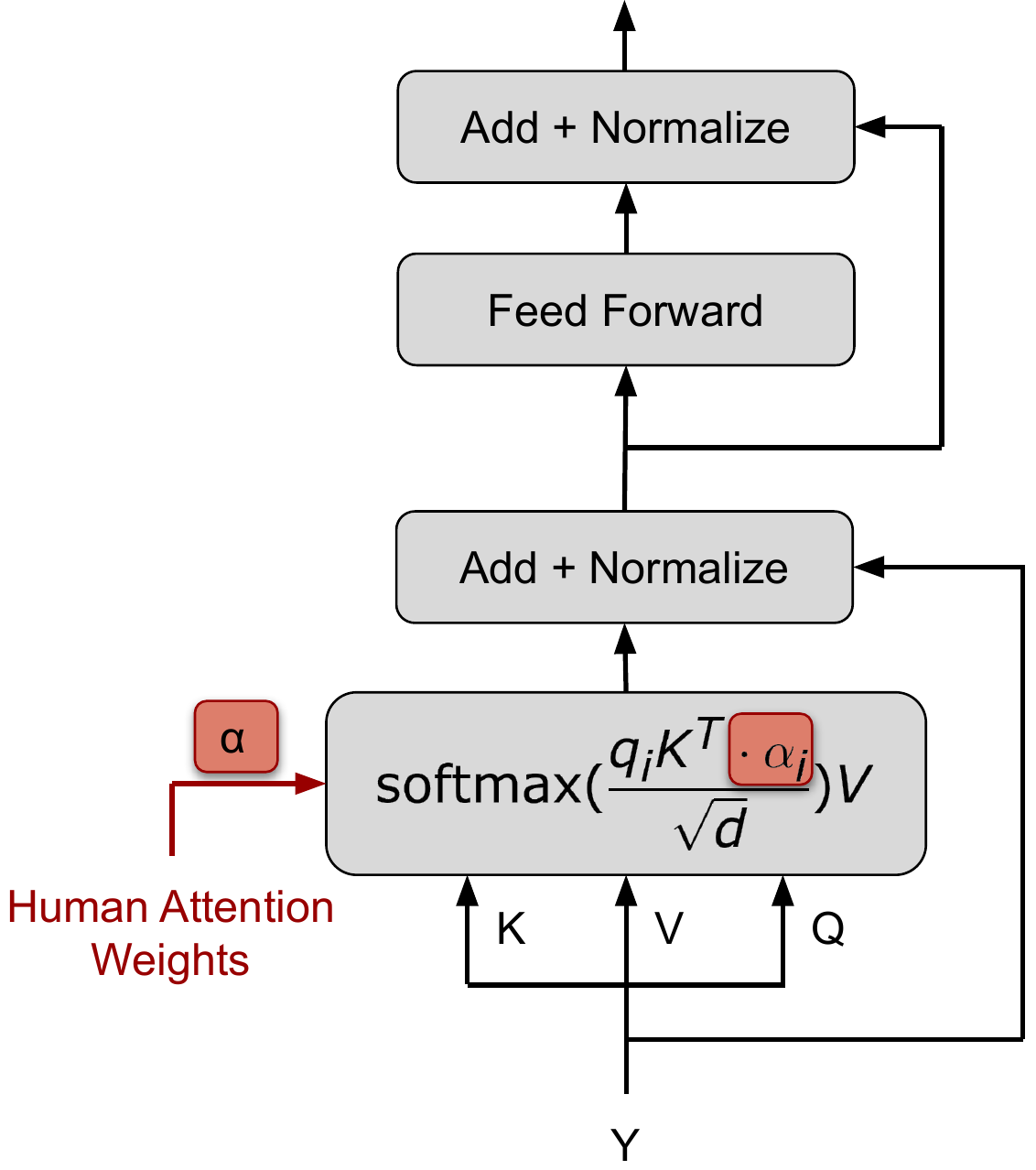}
  \caption{Schematic of a self-attention (SA) layer. The vanilla SA layer is shown in gray, while our human attention integration approach in red.}
  \label{fig:atention_integration_SA}
\end{figure}

A schematic of an SA module is shown in \autoref{fig:atention_integration_SA} in gray. 
It consists of two sub-layers: the multi-headed attention and a feed-forward layer. 
Both are encompassed by a residual connection and layer normalization.
The attention sub-layer projects the input feature embeddings into queries $Q\in \mathbb{R}^{n\times d}$, keys $K\in \mathbb{R}^{n\times d}$, and values $V\in \mathbb{R}^{n\times d}$ with a common hidden dimension $d$. For a query $q$ the attended output is calculated with:
\begin{equation}
A(q,K,V) = \textrm{softmax}(\frac{qK^T}{\sqrt{d}})V
\end{equation}
As in the Transformer~\cite{Vaswani.2017}, this is calculated for multiple queries at once with $QK^T$ and the results of several heads with different projections for $Q$, $K$, $V$ are combined.

The GA module is set up identically to the SA module except the queries and key-value pairs are provided by separate inputs. 
In this way, text features can guide attention on image features. 
Intuitively, the attention layer reconstructs the queries from a linear combination of the values, emphasizing interactions between them. The value space is projected from the input features, which in the GA case is a fusion space between the modalities.

The MCAN encoder stacks multiple layers of SA on text features $Y\in\mathbb{R}^{n_y\times d_y}$ before feeding the result of the last layer into the decoder. The decoder stacks modules with SA on image features $X\in\mathbb{R}^{n_x\times d_x}$ and GA between the encoder result and the SA output. After the last layer, the resulting feature matrices from both encoder and decoder are flattened to obtain the attended features $\tilde{y}\in\mathbb{R}^d$ and $\tilde{x}\in\mathbb{R}^d$ 
and fused by projecting them into the same space and adding them. The VQA task is formulated as classification, so a final projection into the answer dimension and a sigmoid function conclude the network.
\citet{jiang_defense_2020} improved the performance of the original MCAN by replacing the region image features with spatial grid features. We use their model as a baseline for our experiments. 

\paragraph{Human-Like Attention Integration.}
Although the importance of fusing both modalities has been underlined by many previous works and is the driving idea behind co-attention, the integration of external guidance has only been explored in the image domain~ \cite{Gan.2017, qiao_exploring_nodate, Selvaraju.2019, Wu.2019, chen_air_2020, Shrestha.12.04.2020}.

We integrate human-like attention into both the text and image streams of the MCAN base model. For both streams, we use the same basic principle of integration into SA modules (see \autoref{fig:atention_integration_SA}).
We propose a new attention function $A_H$ for the SA layer, multiplying human-like attention weights $\alpha \in \mathbb{R}^{n}$ ($\mathbb{R}^{m}$ for image features) into the attention score. 
For the query $q_i$ corresponding to the $i$-th input feature embedding, we calculate the $i$-th attended embedding:
\begin{equation}
A_H(q,K,V,\alpha) = \textrm{softmax}(\frac{q_iK^T \cdot \alpha_i}{\sqrt{d}})V
\end{equation}

We integrate human-like attention on the question text into the first SA module in the encoder part of MCAN (see \autoref{fig:overall_arch}) and on the image after the first GA module that integrates text and image. 
This early integration is motivated by \citet{Brunner.2020} who investigated the token mixing that occurs in self-attention layers. 
They found that the contribution of the original input token to the embedding at the same position quickly decreases after the first layer, making the integration of re-weighting attention weights less targeted at later layers.
We opted to integrate human-like image attention in the SA module \emph{after} the first GA module (as opposed to before) because this allows text-attention dependent features to interact during integration of human-like attention on the image. To obtain human-like attention scores for questions and images, we make use of domain-specific attention networks that we discuss in the following. 

\paragraph{Text Attention Model.}
For text, we make use of the recently introduced Text Saliency Model (TSM)~\cite{sood_improving_2020} that yields an attention weight for every token in the question.
TSM is pre-trained on synthetic data obtained from a cognitive reading model as well as on real human gaze data.
\citet{sood_improving_2020} proposed a joint training approach in which TSM predictions are integrated into the Luong attention layer~\cite{luong2015effective} of a downstream NLP task and fine-tuned while training for this downstream task.
We follow a similar methodology and fine-tune the TSM while training our VQA network.

\paragraph{Image Attention Model.}
For images, we obtain human-like attention using the state-of-the-art Multi-Duration Saliency (MDS) method~\cite{fosco2020much}.
MDS predicts human attention allocation for viewing durations of 0.5, 3, and 5 seconds.
Because our integration approach requires a single attention map per image, we use the output of MDS for the 3 second viewing duration as suggested by the original authors.
The input images are of different aspect ratios which leads to black borders in the obtained fixed-size MDS maps after re-scaling.
However, the grid features are extracted from images with their original aspect-ratios.
Therefore, to obtain a single attention weight per feature, we remove the borders from the MDS maps, overlay the grid features and sum the pixel values in every grid cell.
The values are then normalized over the total sum to produce a distribution.

\begin{table}[t]
    \centering
    \begin{tabular}{rrr}
    \toprule
    Model & \textit{test-std} & \textit{test-dev} \\
    \midrule
    \methodName (multimodal) & \textbf{73.98\%} & \textbf{73.72\%} \\
    Text only (TSM) & 73.77\% & 73.52\% \\
    Image only (MDS) & 73.67\% & 73.39\% \\
    No Integration & 73.65\% & 73.39\% \\
    \midrule
    Li et al. (2020) & 73.82\% & 73.61\% \\ 
    Jiang et al. (2020) & 72.71\% & 72.59\%\\
    \bottomrule
    \end{tabular}
    \caption{Results showing \textit{test-std} and \textit{test-dev} accuracy scores of our model (trained on \textit{train+val+vg}) and ablated versions over  different datasets. \methodName achieves state-of-the-art on both benchmarks.}
    \label{tab:acc_test_main}
\end{table} 
\paragraph{Implementation Details.}
We trained the network using the basic configuration and hyperparameters of MCAN\_small. The input features were set to $d_y = 512$ and after the change to grid features, $d_x = 2048$. The hidden dimension $d$ inside the Transformer heads was kept at 512. The increased dimensions in the MCAN\_large configuration did not bring performance advantages in our preliminary experiments, so we opted for fewer parameters.
In the added TSM model we set the hidden dimension for both BiLSTM and Transformer heads to 128. We used 4 heads and one layer.
Even including the trainable parameters added by the TSM model, our full \methodName model has significantly fewer parameters than MCAN\_large (\methodName: 58M, MCAN\_large: 203M).
We kept the MCAN Adam Solver and the corresponding learning rate schedule and trained over 12 epochs with batch size 64. The pre-trained TSM model is trained jointly with the MCAN.
Results on \textit{test-std} were obtained after training on \textit{train} and \textit{val} splits and a subset of Visual Genome \textit{vg}, for results on \textit{val} we trained on \textit{train}. 
We trained on single Nvidia Tesla V100-SXM2 GPUs with 32GB RAM. Average runtime was 2.6h for models trained on \textit{train} (36h for convergence) and 5.3h for models trained on \textit{train+val+vg} (68h for convergence).
\section{Experiments}
We used VQAv2\footnote{\url{https://visualqa.org/download.html}}, the balanced version~\cite{Goyal.2017} of the VQA dataset~\cite{antol_vqa_2015}, for all our experiments. 
VQAv2 is among the most popular benchmark datasets in the field and contains 1.1M human-annotated questions on 200K images from MS COCO~\cite{lin2014microsoft}, split into \textit{train}, \textit{val}, and \textit{test-dev}/\textit{test-std} sets.
The annotations for the test splits have been held back for online evaluation of models submitted to the annual challenge.
The standard evaluation metric\footnote{\url{https://visualqa.org/evaluation.html}} is simple overall accuracy, for which agreement with three out of the 10 available annotator answers is considered as achieving an accuracy of 100\%. In practice the machine accuracy is calculated as mean over all 9 out of 10 subsets.

The standard evaluation is misleading because the same overall accuracy can be achieved by answering very different sets of questions. 
Hence, we evaluated accuracy overall for comparison with other work, but also binned ``per question-type" as proposed by \citet{kafle_analysis_2017} to compensate for class imbalance and answer bias skewing the evaluation. 
For this, we used their question types that categorize questions by the task they solve and added a \textit{reading} category for questions that are answered by text on the image. Because they label only about 8\% of VQAv2 \textit{val}, we annotated the full \textit{val} set using a BiLSTM network pre-trained on TDIUC (1.6M samples) and hand-crafted regular expressions (245K samples).
These additional annotations allowed us to assess the performance changes of our model in more detail.

We performed a series of experiments to evaluate the performance of our proposed method.
To shed more light on the importance of multimodal integration, we first compared different ablated versions of our method on \textit{test-dev} and \textit{test-std}.
Specifically, we compared multimodal with text-only, image-only, and integration of human-like attention.
Afterwards, we evaluated performance of the multimodal method when integrating human-like attention at different layers of the Transformer network.
Finally, we evaluated more fine-grained performance values of the multimodal, unimodal, and no attention integration method for the different question types.
For all of these experiments, we report accuracy on \textit{test-std} with training on the union of \textit{train}, \textit{val} and \textit{vg} sets.

\begin{figure}[ht]
  \centering
  \includegraphics[width=0.9\columnwidth]{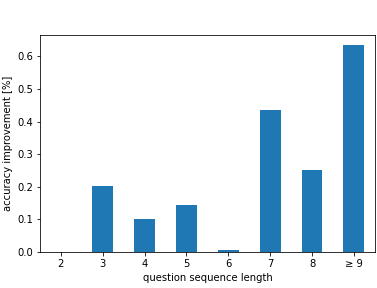}
  \caption{Performance improvements of our multimodal integration method (\methodName) relative to the baseline (\textit{No Integration}), depending on the question length. Results show a significant increase in accuracy for longer questions, in particular when questions have seven or more tokens. 
  }
  \label{fig:sequence_length_accuracy}
\end{figure}

\section{Results and Discussion}

\begin{table}[t]
      \centering
    \begin{tabular}{rrr}
    \toprule
    TSM & MDS & \textit{test-std} \\
    \midrule
    1 & 2 &  \textbf{(ours) 73.98\%} \\
    2 & 2 & 73.64\% \\ 
    1, 3, 5 & 2 & 73.73\% 
    \\
    1 & 1--6 & 71.55\% 
    \\
    1--3 & 2 & 73.49\% \\ 
    1--6 & 2 & 73.73\% \\ 
    1--6 & 2--6 & 73.50\% \\ 
    \bottomrule
    \end{tabular}
    \caption{Layer-wise integration ablation study results, on \textit{test-std}. We integrate human-like attention at different layer combinations. TSM question attention weights are integrated into encoder SA modules, MDS image attention weights into decoder SA modules. 
    }
    \label{tab:integration_exps_test_acc}
\end{table}

\begin{table*}[]
    \centering
    \begin{tabular}{rr|rrrrrr}
    \toprule
    Question type & Bin Size & No Integration & text-only & image-only & \textbf{\methodName} \\
    \midrule
reading & 31\,K &\textbf{42.46} & 42.28 & 42.40 & 42.30  \\
activity recognition & 15\,K & 74.55 & 74.72 & 74.59 & \textbf{75.01} \\
positional reasoning & 26\,K & 61.74 & 61.97 & 61.85 & \textbf{62.01} \\
object recognition & 28\,K & 82.59 & 82.50 & 82.49 & \textbf{82.68} \\
counting & 24\,K & 59.77 & 59.70 & 59.44 & \textbf{59.82} \\
object presence & 17\,K & 86.45 & 86.47 & \textbf{86.59} & 86.57 \\
scene recognition & 15\,K & 79.19 & 79.10 & \textbf{79.20} & 79.19 \\
sentiment understanding & 14\,K & 83.59 & 83.77 & 83.53 & \textbf{83.92} \\
color & 25\,K & \textbf{80.56} & 80.52 & 80.31 & \textbf{80.56} \\
attribute & 4\,K & 69.36 & 69.09 & 69.37 & \textbf{69.65} \\
utility affordance & 11\,K & 66.33 & 66.40 & \textbf{66.64} & 66.42 \\
sport recognition & 6\,K & 85.39 & 85.38 & \textbf{85.93} & 85.60  \\
\midrule
 Overall VQAv2 \textit{val} Accuracy: & & 70.06 & 70.09 & 70.03 & \textbf{70.28*}\\
    \bottomrule
    \end{tabular}
    \caption{Performance on VQAv2 \textit{val} 
    split in terms of per-question-type accuracy of the proposed multimodal integration method (\methodName) and the unimodal ablations \textit{text-only} or \textit{image-only} and no integration of human attention (\textit{No Integration}). Because the online evaluation of VQAv2 only returns overall accuracy, we cannot obtain fine-grained accuracy for \textit{test-std} or \textit{test-dev}. A star indicates statistically significant $p$ at $p < 0.05$.}
    \label{tab:fine_grained_val_acc}
\end{table*}

\begin{figure*}[ht]
  \centering
  \hspace{-1.3cm}
  \includegraphics[width=0.75\textwidth]{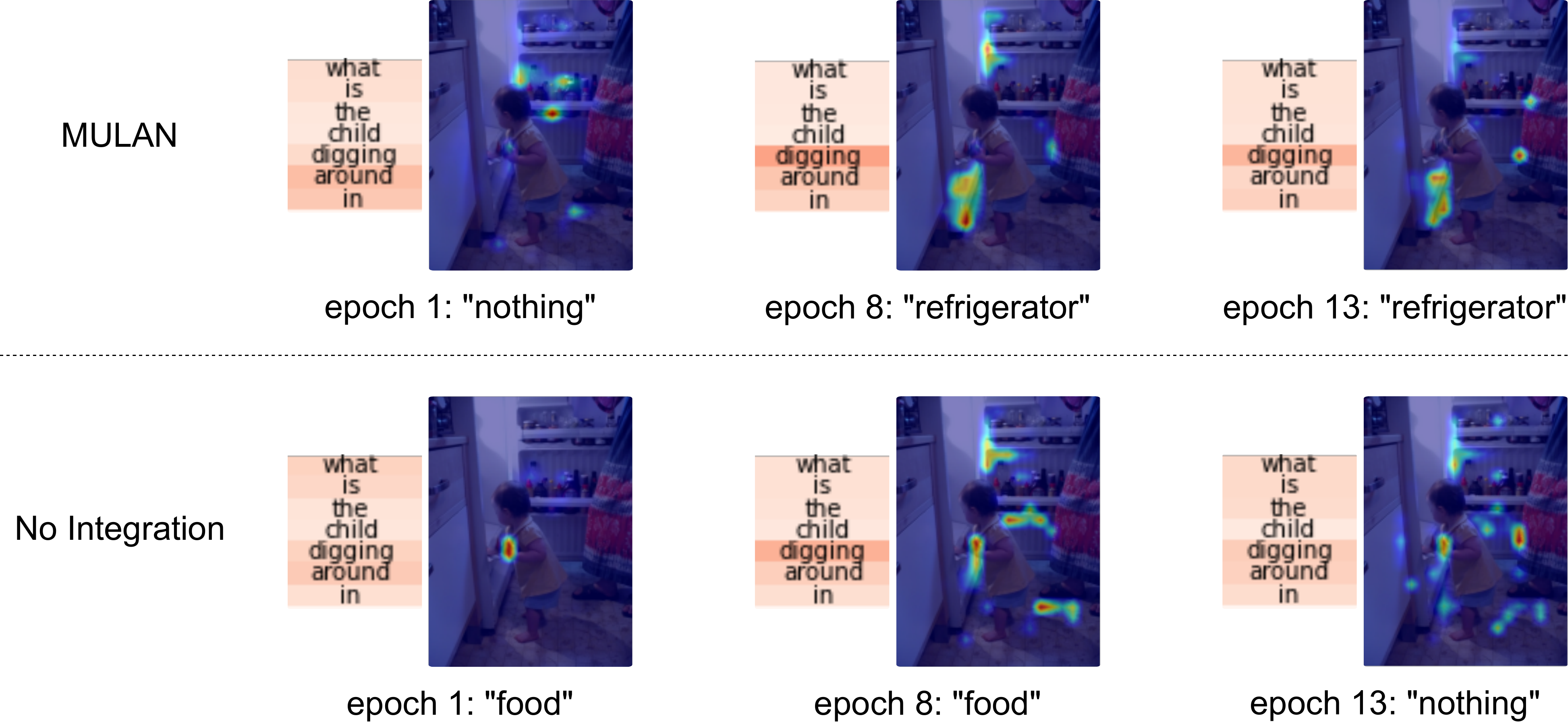}
  \caption{
  Visualization of the weights in the attention reduction modules for text and image features. We compared \methodName and the baseline model (No Integration) at epochs 1, 8, and 13. The input question was ``What is the child digging around in?", the correct answer is ``fridge''. Classification outputs are given for each considered epoch.
  }
  \label{fig:attention_vis}
\end{figure*}

\paragraph{Overall Performance.} \autoref{tab:acc_test_main} shows results of our \methodName model along with current state-of-the-art approaches and ablations of our method.
Our method outperforms the current state of the art~\cite{li_oscar_2020}, reaching accuracy scores of 73.98\% on \emph{test-std} and 73.72\% on \emph{test-dev}, as compared to 73.82\% and 73.61\%. 
Our model also uses approximately 80\% less trainable parameters than ~\citet{li_oscar_2020}.
Notably, we observe a systematic increase in performance as a result of human-like attention integration.
Our base model without any integration reaches 73.65\% on \emph{test-std}.
While the integration of human-like attention on only images (73.67\%) or text (73.77\% on \emph{test-std}) can lead to an increase in performance, our full \methodName model employing multimodal integration is the best performing approach. This further underlines the importance of jointly integrating human-like attention on both text and images for the VQA task, which is inherently multimodal.

\paragraph{Layer-Wise Integration Experiments.}
We evaluated the integration of human-like attention on questions and images for different layers in the MCAN encoder-decoder architecture (see \autoref{tab:integration_exps_test_acc}). We investigated the integration of TSM outputs into different SA layers of the encoder, and the integration of MDS outputs into different SA modules of the decoder.
Among all investigated combinations, the initial integration into the first layer of the encoder and the second layer of the decoder performed best (73.98\% accuracy).
Integrating TSM predictions into the second encoder layer decreased the overall accuracy to 73.64\%, which is in line with the reasoning discussed in~\citet{Brunner.2020}, where with layer depth feature embeddings are increasingly mixed and therefore less attributable to the input word token at the same position. The TSM predicts attention weights for specific word tokens.
We further investigated the integration of TSM and MDS predictions at multiple layers in the MCAN architecture. However all options resulted in decreased performance in comparison to \methodName.
Our results indicate that early integration of human-like attention at a single point for both text and image is optimal for the VQA task.

\paragraph{Category-Specific Performance.} 
To obtain a deeper understanding of our improvements over baseline approaches, we categorized question types into 12 fine-grained bins, similarly to \citet{kafle_analysis_2017}.
\autoref{tab:fine_grained_val_acc} shows a detailed breakdown of accuracy results by category type. 
We used the validation set, rather than the test set, since we needed access to the ground truth annotations to calculate the per-category accuracy. For the same reason, we can only perform the paired t-test on the full validation set. As can be seen, all ablated models obtain inferior performance to our full model on the validation set (statistically significant at the 0.05 level).
For most categories, \methodName achieves the highest accuracy. 
Moreover, in comparison to the baseline, our method is the best performing one in 10 out of 12 categories with especially clear improvements in activity recognition and sentiment analysis categories. 
\methodName expectedly reduces accuracy on reading questions that other models can most likely only answer by bias exploitation, and improves on small bins like attribute. 
The distances between the models are small in absolute terms, but given the vastly different bin sizes the relative improvements are large.
This underlines the robustness of improvements with human-like attention integration and, in particular, multimodal integration.

\paragraph{Sequence-Length Analysis.}
Previous works have shown that VQA models often converge to the answer after processing only the first words of a question, a behavior that has been characterized as ``jumping to conclusions''~\cite{agrawal_analyzing_2016}. 
Human-like attention integration might be useful to combat this effect as the TSM was previously shown to successfully predict human-like attention distributions across all salient words in the input sequence~\cite{sood_improving_2020}. 
As this effect might be especially pronounced for longer questions, 
we investigated whether human-like attention integration in \methodName can especially improve on those questions~\cite{agrawal_analyzing_2016}.
\autoref{fig:sequence_length_accuracy} shows the results of an evaluation where we analyzed the improvements of our system relative to the baseline model, depending on the question length. We find that while \methodName improves for all questions independent of their length, its advantage is especially significant for questions that contain seven tokens or more (relative improvements of 0.3\% or more), indicating that \methodName can improve upon the above-described challenge.

\begin{figure}[ht]
  \centering
  \includegraphics[width=0.9\columnwidth]{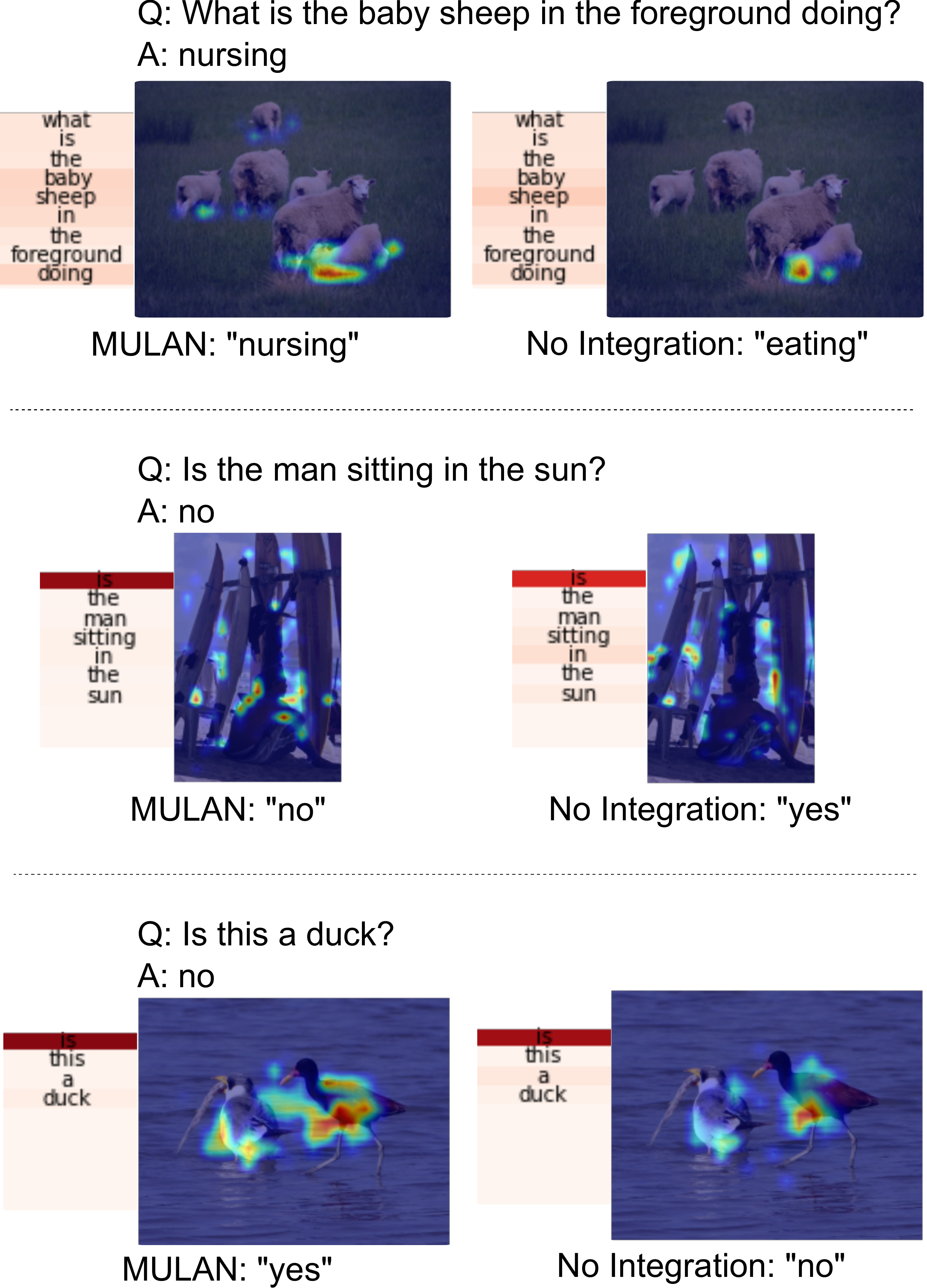}
  \caption{
  Visualization of attention distributions for \methodName and the baseline (No Integration). The upper examples show improvements of \methodName over the baseline, while the bottom shows a failure case.
}
  \label{fig:attention_vis_stationary}
\end{figure}

\paragraph{Attention Visualizations.} 
To further investigate \methodName's ability to answer longer questions than the baseline model, we visualized the attention weights in the attention reduction modules after the encoder/decoder, which merge the text and image features to one attended feature vector each.
These weights represent the final impact of the transformed features.
\autoref{fig:attention_vis} shows examples from the validation set with the corresponding predictions comparing our method to the baseline at epochs 1, 8 and 13. The input question was ``What is the child digging around in?'' and the correct answer is ``fridge''. Our method it able to correctly predict that the child is digging in the fridge as opposed to the baseline that outputs ``nothing''.
\methodName focuses on both the token ``digging'' as well as the location, which is in front of the child. 
In contrast, the attention of the baseline model is more spread out, failing to focus on the relevant cues. Interestingly, the focus of attention in the baseline evolved over several epochs of training, unlike \methodName which quickly converged to a stable attention distribution. This indicates that initial human-like attention maps on the image are indeed adapted using attention-based information extracted from the question text.
\autoref{fig:attention_vis_stationary} shows three additional examples of our method compared to the baseline from the final epoch. The top and middle examples show how our method is able to correctly answer the question, while the baseline fails. The bottom example shows an error case of our method.

\section{Conclusion}

In this paper, we propose the first method for multimodal integration of human-like attention on both image and text for visual question answering.
Our Multimodal Human-like Attention Network (\methodName) method integrates state-of-the-art text and image saliency models into neural self-attention layers by modifying attention scoring functions of transformer-based self-attention modules.
Evaluations on the challenging VQAv2 dataset show that our method not only achieves state-of-the-art performance (73.98\% on \textit{test-std} and 73.72\% on \textit{test-dev}) but also does so with fewer trainable parameters than current models.
As such, our work provides further evidence for the potential of integrating human-like attention as a supervisory signal in neural attention mechanisms.

\clearpage
\bibliographystyle{acl_natbib}
\bibliography{references}

\end{document}